\documentclass[DTMColor]{ROB-New}

\usepackage{float}
\usepackage{subfigure}
\usepackage{booktabs} 
\usepackage{multirow} 
\usepackage{graphicx} 
\AtBeginDocument{%
  \hypersetup{colorlinks=true,linkcolor=blue,citecolor=blue,urlcolor=blue}%
}

\begin{document}

\authormark{Chen et al.}

\articletype{RESEARCH ARTICLE}

\jnlPage{1}{7}
\jyear{2026}
\jdoi{10.1017/xxxxx}

\title{HierKick: Hierarchical Reinforcement Learning for Vision-Guided Soccer Robot Control}

\author[1,2]{Yizhi Chen}
\author[2,3]{Zheng Zhang}
\author[2,3]{Zhanxiang Cao}
\author[4]{Yihe Chen}
\author[1,2]{Shengcheng Fu}
\author[2,3]{Liyun Yan}
\author[2,3]{Yang Zhang}
\author[2,5]{Jiali Liu}
\author[2,3]{Haoyang Li}
\author*[2,3]{Yue Gao\hyperlink{corr}{*}}

\address[1]{Tongji University, Shanghai, China}
\address[2]{Shanghai Innovation Institute, Shanghai, China}
\address[3]{Shanghai Jiao Tong University, Shanghai, China}
\address[4]{Jiangxi University of Finance and Economics, China}
\address[5]{East China Normal University, Shanghai, China}
\address{\hypertarget{corr}{*}Corresponding author. \email{yuegao@sjtu.edu.cn}}



\abstract{Controlling soccer robots involves multi-time-scale decision-making, which requires balancing long-term tactical planning and short-term motion execution. Traditional end-to-end reinforcement learning (RL) methods face challenges in complex dynamic environments. This paper proposes HierKick, a vision-guided soccer robot control framework based on dual-frequency hierarchical RL. The framework adopts a hierarchical control architecture featuring a 5 Hz high-level policy that integrates YOLOv8 for real-time detection and selects tasks via a coach model, and a pre-trained 50 Hz low-level controller for precise joint control. Through this architecture, the framework achieves the four steps of approaching, aligning, dribbling, and kicking. Experimental results show that the success rates of this framework are 95.2\% in IsaacGym, 89.8\% in Mujoco, and 80\% in the real world. HierKick provides an effective hierarchical paradigm for robot control in complex environments, extendable to multi-time-scale tasks, with its modular design and skill reuse offering a new path for intelligent robot control.}

\maketitle

\section{INTRODUCTION}

Recent research on humanoid robots has advanced from achieving stable walking\cite{zhang2017deep,song2021deep,liskustyawati2024effect} to completing complex tasks in real-world scenarios\cite{luo2025precise,lin2025sim,gu2025humanoid}. This progress parallels the sophisticated neural regulatory mechanisms in biological evolution. For instance, when an athlete chases a rolling soccer ball, the brain acts as a strategic command center, making rapid decisions such as whether to dribble past an opponent or pass the ball for cooperation. Meanwhile, the cerebellum functions as a precision actuator, constantly adjusting muscle tension to maintain balance and calibrate the kicking angle. The synergy between high-level decision-making and low-level execution exemplifies how organisms tackle multi-time-scale tasks.

The football task serves as an excellent long-sequence execution verification task\cite{aguiar2017regularity}. Given the unpredictable trajectory of the ball, the strategy must be robust. Visual perception, susceptible to occlusions and lighting changes, prevents the robot from acquiring complete global information. As a result, the robot must plan tactics for approaching the ball\cite{giordano2019dribbling}, dribbling forward, and aiming or shooting at the high level, while simultaneously maintaining balance, adjusting step length, and controlling kicking force at the low-frequency level. A disconnect between these two levels will directly lead to task failure.

\begin{figure}
    \centering
    \includegraphics[width=0.8\linewidth]{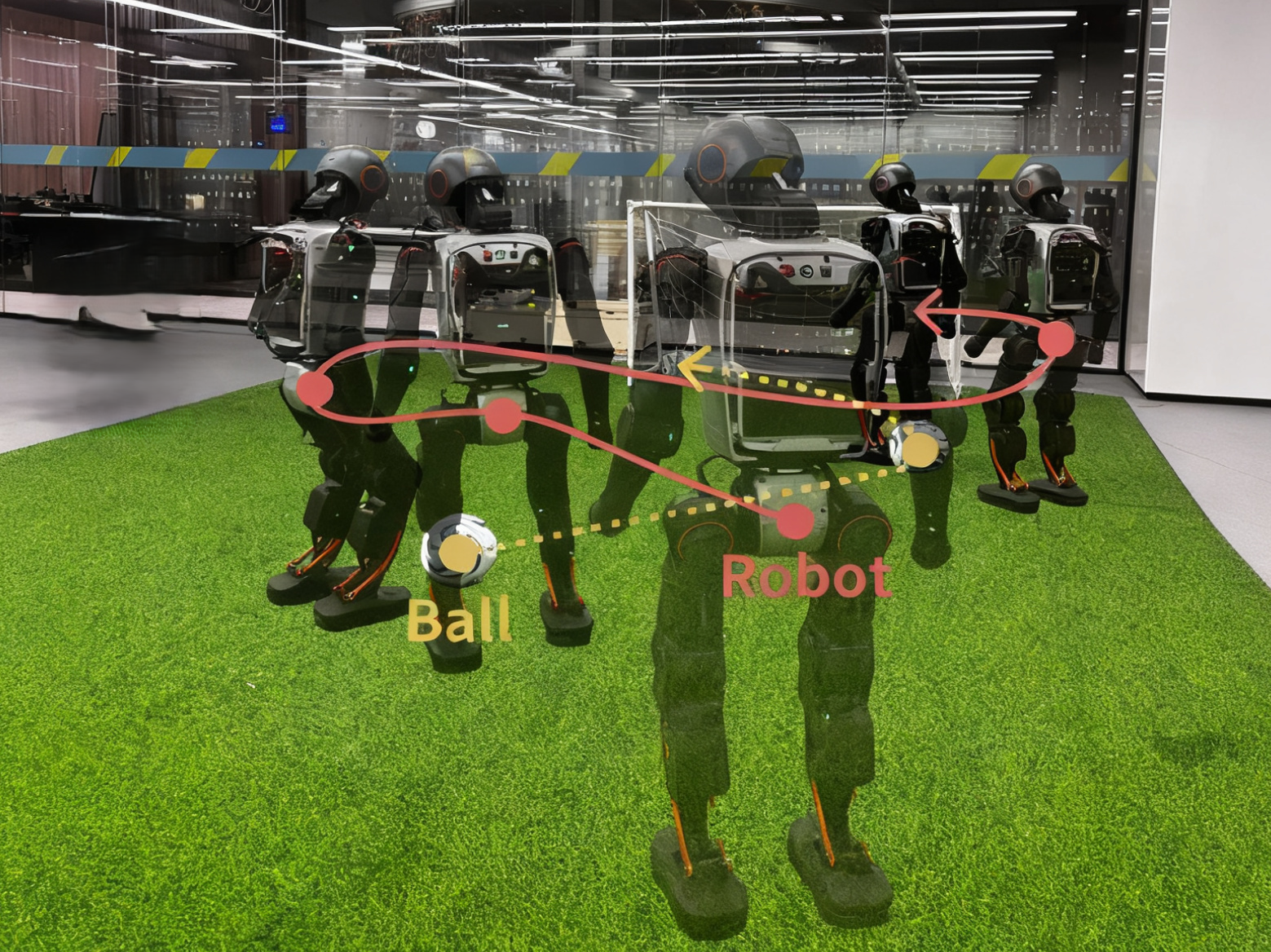}
    \caption{\textbf{Robot Trajectory Diagram}. This diagram depicts the trajectory of the robot on the football field. Guided by the Coach model and based on the HierKick framework, the robot executes a series of actions, including Approach, Alignment, Dribble, and Shoot.}
    \label{fig:shoot}
\end{figure}

Current technology faces several limitations: reward sparsity leads to convergence difficulties\cite{devidze2022exploration,hu2025towards}, strategies lack stability, and control delays are high, making it difficult to meet low-level action requirements\cite{abadia2021cerebellar}. Additionally, large-scale manual intervention is often necessary, and adapting to dynamic environmental changes is challenging\cite{pei2021improved}. To address these issues, this paper proposes HierKick, a dual-frequency hierarchical reinforcement learning framework for visually guided soccer robots. The core idea is to balance multiple time-scale demands by decoupling and synergizing ``tactical decision-making - motion execution,'' integrating pre-trained skills, and utilizing multi-modal perception to enhance robustness.

We propose the HierKick framework, which incorporates hierarchical frequency settings. Specifically, a coach model is trained to guide the robot in selecting appropriate speeds for different tasks. The coach model outputs acceleration, which is integrated to generate the robot's velocity command for each high-level time step. This design allows HierKick to achieve a 95.2\% success rate in the approaching--dribbling--shooting composite task in a simulated environment and an 80\% success rate in real-world deployment. These results validate the framework's effectiveness and transferability. Comparative and ablation experiments further demonstrate that dual-frequency hierarchical control enhances strategy stability and real-time performance, while the reuse of pre-trained skills accelerates training and improves convergence. Additionally, large-scale parallel training and noise modeling increase robustness and sample efficiency. At a 50 Hz control frequency, the system achieves an end-to-end control latency of approximately 20 ms, meeting real-time control requirements.

The main contributions of this paper are as follows:
\begin{itemize}
    \item We propose HierKick, a dual-frequency hierarchical reinforcement learning framework with hierarchical frequency settings that enable adaptive speed control for the robot.
    \item The framework incorporates a coach model trained through curriculum learning and a hierarchical reward function. This model evaluates on-field situations and outputs acceleration signals, which are integrated to generate velocity commands for each high-level time step, allowing the robot to gradually master various skills.
    \item We successfully implement the hierarchical control strategy on the Booster T1 physical humanoid robot, achieving an end-to-end control latency of approximately 20 ms at a 50 Hz control frequency, meeting the practical requirements for real-time control.
\end{itemize}

\section{RELATED WORK}

\subsection{Dynamic Locomotion on Humanoid Robots with RL}
In recent years, reinforcement learning (RL) has achieved remarkable success in controlling the dynamic locomotion of humanoid robots\cite{smith2023learning,hundt2024towards,nguyen2025mastering}. Numerous studies have applied model-free RL techniques\cite{morimoto2021model} to enable diverse and agile motor skills. By utilizing multistage training schemes, the robustness of robots in tasks such as jumping\cite{chae2022agile} and rapid recovery\cite{liu2023robot} has been significantly improved. Additionally, several works have integrated vision systems to help robots navigate complex terrains\cite{hu2021sim,pei2021improved,lee2024learning}. However, these studies primarily focus on the locomotion aspects of humanoid robots. In contrast, our method, HierKick, addresses dynamic soccer tasks, which require not only stable locomotion but also precise interaction with external objects.

\subsection{Learning-Based Robot Soccer Skills}
Robot soccer has long been a benchmark task for legged robot locomotion and coordination, particularly within the RoboCup competition series\cite{beukman2024robocupgym}. While early RoboCup participants often relied on rule-based or scripted control strategies\cite{yi2016hierarchical}, there has been an increasing shift toward learning-based methods. Quadruped robots, in particular, have demonstrated impressive skills in dribbling and goalkeeping using RL\cite{abreu2025designing}. However, approaches for bipedal robots\cite{da2021deep} have typically been limited to discrete kicking actions, lacking continuous control over the ball. Some end-to-end learning methods have achieved agile control in simulations, but they often struggle with the challenges of multi-time-scale decision-making\cite{tirumala2024learning}, resulting in difficulties in policy convergence and high control latency.

To address these issues, our proposed HierKick method adopts Hierarchical Reinforcement Learning (HRL)\cite{wang2023hrl}, an effective paradigm that simplifies learning by decomposing tasks into a high-level coach policy and a low-level policy. Unlike many conventional HRL approaches, where simultaneous learning of both policies can cause training instability\cite{dong2021motion}, HierKick utilizes a pre-trained low-level motion controller. This design treats the low-level policy as a fixed motion generator, mitigating the cold-start problem, improving training stability and efficiency, and allowing the high-level policy to focus on long-term task planning.

\section{METHOD}
\begin{figure}[htbp]
  \centering
  \includegraphics[width=1.0\linewidth]{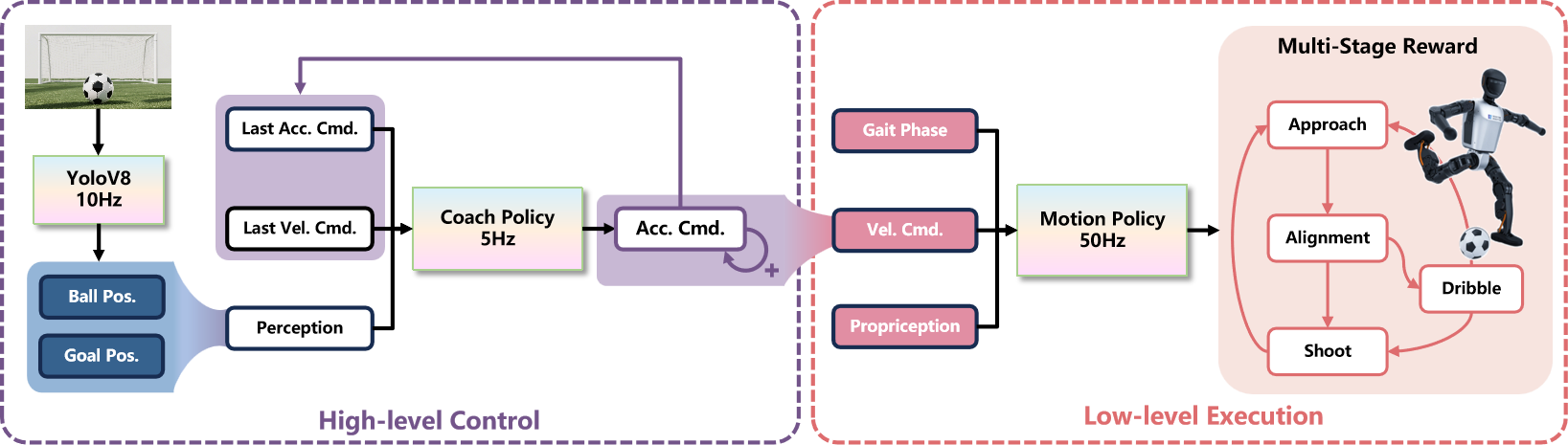}
  \caption{\textbf{Hierarchical control framework of HierKick}. This framework consists of three interconnected modules: (1) \textit{Perception}, where a YOLOv8 detector running at 10 Hz processes visual input to determine the positions of the soccer ball and goal (2) \textit{High-level control}, featuring a ``Coach Policy'' operating at 5 Hz, which generates acceleration commands through PPO reinforcement learning by fusing perception results and historical control data (last acceleration and velocity commands) (3) \textit{Low-level execution}, where a ``Motion Policy'' running at 50 Hz takes inputs like velocity commands, gait phase, and proprioception to execute sequential soccer actions (Approach, Alignment, Dribble, Shoot), with training guided by a multi-stage reward mechanism.}
  \label{fig:hierkick-framework}
\end{figure}

The HierKick framework for humanoid robot dynamic soccer tasks is divided into two main components (see Fig.~\ref{fig:hierkick-framework}): \textit{High-level Control} and \textit{Low-level Execution}, enabling time-scale decoupling between tactical decision-making and motion implementation. In the high-level control module, which operates at 5 Hz, YOLOv8 provides real-time visual perception of the ball and goal. The Coach Policy integrates this perceptual data with historical acceleration and velocity commands to generate acceleration commands, enabling adaptive speed regulation for strategic task planning. Additionally, we propose a multi-stage reward mechanism that plays a key role in facilitating the learning process. This mechanism is designed for four tasks: Approach, Alignment, Dribble, and Shoot. To prevent erratic control and ensure smooth transitions between commands, a regularization term is incorporated into the reward function using delta logic. This integrated design avoids conflicting objectives, enables sequential skill accumulation, and ensures context-appropriate motion.

\subsection{Problem Statement}
We model the control of a humanoidal soccer robot as a hierarchical Markov Decision Process (MDP)\cite{bolshakov2023hierarchical} with two distinct decision-making levels. The high-level MDP is defined as $(S, A, f, r, S, \gamma)$, where $S$ represents the high-level state space, $A$ is the high-level action space, $f$ is the high-level system dynamics, $r$ is the high-level reward function, $S$ is the initial high-level state, and $\gamma$ is the discount factor.

Similarly, the low-level MDP\cite{park2023predictable} is defined as $(S, A, f, r, S, \gamma)$. The objective of the HierKick approach is to learn a high-level policy $\pi$ that maximizes the expected discounted return $G_t = \mathbb{E}_{\pi}\left[\sum_{t=0}^{T-1} \gamma^t r^t\right]$, while utilizing a pre-trained low-level policy $\pi$ for locomotion execution. We employ Proximal Policy Optimization (PPO) \cite{schulman2017proximal} with an asymmetric actor-critic architecture, using privileged information during training and relying on partial observations during deployment. The low-level policy is based on conventional locomotion training methods, so we do not elaborate on it further in this paper.

\subsection{High-Level Coach Policy}
The high-level coach policy operates at a frequency of once every 10 simulation steps, which corresponds to 200 ms intervals. This frequency aligns with the timescale of human tactical decision-making, helping to avoid over-reactive behaviors that could destabilize the soccer play. At each timestep, the policy receives soccer-specific observations and outputs velocity increment commands to guide the robot's actions.

\subsubsection{Observation Space}
The high-level observation space is composed of three main categories: soccer-specific perception, the robot's state, and command history. These components collectively inform the decision-making process. The observation space is formally defined as follows:
\[
\mathbf{o_{\text{high-level}}} = \left[ \mathbf{p}_{\text{robot-ball}}, \,  \mathbf{p}_{\text{robot-goal}}, \, d_{\text{ball}}, \, d_{\text{goal}}, \, \mathbf{v}_{\text{robot}}, \, \mathbf{c}_{\text{prev}} \right]
\]
Here, $\mathbf{v}_{\text{robot}}$ represents the robot's linear velocities along the x-axis and y-axis, as well as its angular velocity around the z-axis. The term $\mathbf{p}_{\text{robot-goal}}$ refers to a vector pointing from the robot's base to the center of the goal, which is computed using field landmarks detected by the robot's vision system. $d_{\text{ball}}$ and $d_{\text{goal}}$ represent the straight-line distances from the robot to the ball and the goal, respectively. Finally, $\mathbf{c}_{\text{prev}}$ denotes the velocity increment command issued in the previous timestep, which helps maintain continuity in the robot's movement.

\subsubsection{Action Space}
The high-level action space is a 3-dimensional continuous vector of velocity increments $\Delta \mathbf{v} = [\Delta v_x, \Delta v_y, \Delta \omega_z]$, where $\Delta v_x$, $\Delta v_y$, and $\Delta \omega_z$ denote the incremental changes in forward/backward linear velocity (x-axis), lateral linear velocity (y-axis), and yaw angular velocity (z-axis), respectively. These increments are bounded to ensure safe robot operation: $\Delta v_x \in [-0.2,0.2]$ m/s, $\Delta v_y \in [-0.1,0.1]$ m/s, and $\Delta \omega_z \in [-0.1,0.1]$ rad/s. The velocity command is updated recursively to maintain smooth motion:
\[
\mathbf{v}_{\text{cmd}}(t) := \mathbf{v}_{\text{cmd}}(t-1) + \Delta \mathbf{v}(t)
\]
This formulation allows gradual speed adjustment and preserves temporal continuity of the robot's movement.

\subsection{Low-Level Locomotion Controller}
The low-level locomotion controller leverages a pre-trained kick model, which demonstrates robust walking and turning capabilities. It receives velocity commands from the high-level coach policy and generates joint position targets at a frequency of 50 Hz. This frequency ensures precise control over the robot's locomotion, enabling stable and responsive movements during dynamic tasks.

\subsubsection{Observation Space}
The low-level observation space focuses on capturing the real-time dynamic states of the robot's body and motion, providing critical information for maintaining stability and achieving accurate locomotion control. This observation space is 47-dimensional, formally defined as:
\[
\mathbf{o}_{\text{low-level}} = \left[ \mathbf{g}, \, \boldsymbol{\omega}, \, \mathbf{v}_{\text{cmd}}, \, \cos(2\pi\phi), \, \sin(2\pi\phi), \, \mathbf{q}, \, \dot{\mathbf{q}}, \, \mathbf{a}_{\text{prev}} \right]
\]
The 47-dimensional space integrates kinematic, dynamic, and temporal data, enabling the low-level controller to generate stable and responsive joint targets at 50 Hz. By incorporating high-level velocity commands and maintaining precise tracking, the system ensures dynamic adaptability and stability during locomotion.

\subsubsection{Action Space}
The low-level action space is a 12-dimensional continuous space that directly governs the target joint positions of the humanoid's legs. These 12 dimensions correspond to the six degrees of freedom per leg of the humanoid robot. The joint position targets are processed by a PD controller \cite{santibanez2001pd}, which transforms the target positions into actual joint drive signals. This mechanism ensures accurate tracking of the desired positions, mitigates motion oscillations, and maintains the stability of the robot's locomotion throughout task execution.

\subsection{Multi-Stage Reward Mechanism}
To address the conflicting behavioral priorities across different phases of the soccer task, a set of distance-adaptive reward functions is designed. These functions dynamically adjust based on the robot-ball distance, \(d_{\text{ball}}\), ensuring that the high-level policy prioritizes phase-specific objectives while remaining aligned with the overarching task goals.

Building on the principles of curriculum learning \cite{portelas2020automatic} and Finite State Machines (FSM) \cite{liu2022pm} in robotics, which decompose complex tasks into manageable subtasks, a multistage learning strategy is proposed. This strategy adapts reward priorities according to the robot-ball distance, enabling the robot to sequentially acquire phase-appropriate skills without compromising task coherence. Additionally, this approach mitigates the common issue of policies becoming trapped in local optima.

The complete hierarchical reward function integrates all stage-specific components and is expressed as:
\[
r_{\text{total}} = \mathbf{w} \cdot \mathbf{r}
\]
where \(\mathbf{w} = [\lambda_1, \lambda_2, \lambda_3, \lambda_4, \lambda_5]^\top\) is the weight vector, reflecting the relative importance of each phase. More complex skills are assigned higher weights. The vector \(\mathbf{r} = \left[ r_{\text{approach}}, \, r_{\text{alignment}}, \, r_{\text{dribble}}, \, r_{\text{shoot}}, \, r_{\text{delta}} \right]\) represents the rewards for each phase, with \(r_{\text{delta}}\) acting as a regularization term.

\subsubsection{Stage 1 Approach Phase Reward}
When the robot is distant from the ball, the primary objective is efficient, directional locomotion towards the ball. The approach reward prioritizes movement aligned with the ball's direction while minimizing detours. It is defined as:
\[
r_{\text{approach}} = \left( \mathbf{v}_{\text{robot}} \cdot \hat{\mathbf{v}}_{\text{robot-ball}} \right) \cdot M_{\text{approach}}(d)
\]
In this expression, \(\mathbf{v}_{\text{robot}}\) represents the high-level velocity command, and \(\hat{\mathbf{v}}_{\text{robot-ball}}\) is the unit vector pointing from the robot to the ball, ensuring the reward encourages movement toward the ball. \(M_{\text{approach}}(d)\) is a binary stage mask function that activates the reward only during the approach phase, defined as:
\[
M_{\text{approach}}(d) = \begin{cases}
1 & \text{if } d > u_1 \\
0 & \text{otherwise}
\end{cases}
\]
where \(u_1\) represents the threshold for ``far distance'' between the robot and the ball, ensuring that this reward is only active when the robot needs to approach the ball. A positive \(r_{\text{approach}}\) is awarded when \(\mathbf{v}_{\text{robot}}\) aligns with \(\hat{\mathbf{v}}_{\text{robot-ball}}\), encouraging a fast, direct approach.

\subsubsection{Stage 2 Alignment Phase Reward}
As the robot approaches the ball, the focus shifts to positioning the robot for optimal shooting angles. The goal is to align the robot's body so that the ``robot-ball-goal'' line is as straight as possible, minimizing the likelihood of off-target shots. The alignment reward balances angular accuracy with motion smoothness, as follows:
\[
r_{\text{alignment}} = \left[ \cos(\theta_{\text{rbg}}) - \beta \cdot |\Delta \omega_{\text{desired}} - \Delta \omega_{\text{actual}}| \right] \cdot M_{\text{alignment}}(d)
\]
Here, \(\theta_{\text{rbg}}\) is the angle between the unit vector from the robot to the ball and the unit vector from the robot to the goal. The term \(\cos(\theta_{\text{rbg}})\) ranges from 1 (perfect alignment) to -1 (opposite direction), rewarding straight-line positioning. \(\Delta \omega_{\text{desired}}\) is the desired angular velocity for alignment, and \(\Delta \omega_{\text{actual}}\) is the commanded angular velocity. The term \(|\Delta \omega_{\text{desired}} - \Delta \omega_{\text{actual}}|\) penalizes misalignment due to imprecise turning. The parameter \(\beta = 0.5\) balances alignment accuracy with motion smoothness. \(M_{\text{alignment}}(d)\) is the phase mask for the alignment phase, defined as:
\[
M_{\text{alignment}}(d) = \begin{cases}
1 & \text{if } u_2 < d \leq u_1 \\
0 & \text{otherwise}
\end{cases}
\]
where \(u_2\) and \(u_1\) denote the thresholds for ``relatively close'' and ``far'' distances from the ball, respectively. The reward activates when the robot needs to align with the ball.

\subsubsection{Stage 3 Dribbling Phase Reward}
When the robot is close to the ball and distant from the goal, the focus shifts to gently maneuvering the ball toward the goal, avoiding excessive speed that could destabilize the robot or cause the ball to be knocked away. The dribbling reward combines goal-directed movement and speed regulation:
\[
r_{\text{dribble}} = \mathbf{v}_{\text{robot}} \cdot \hat{\mathbf{v}}_{\text{ball-goal}} \cdot M_{\text{dribble}}(d_1,d_2)
\]
Here, \(\hat{\mathbf{v}}_{\text{ball-goal}}\) is the unit vector pointing from the robot to the goal, rewarding movement in the goal direction. \(M_{\text{dribble}}(d_1,d_2)\) is the phase mask for the dribbling phase, defined as:
\[
M_{\text{dribble}}(d_1,d_2) = \begin{cases}
1 & \text{if } 0 < d_1 \leq u_2, \, d_2 > u_3 \\
0 & \text{otherwise}
\end{cases}
\]
where \(d_1\) and \(d_2\) represent the relative distances between the robot and the ball, and between the ball and the goal, respectively. The robot is rewarded for dribbling the ball towards the goal when the conditions are met.

\subsubsection{Stage 4 Shooting Phase Reward}
In the shooting phase, the reward encourages both alignment with the goal and high-speed movement for powerful shots:
\[
r_{\text{shoot}} = \left[ \hat{\mathbf{v}} \cdot \hat{\mathbf{v}}_{\text{ball-goal}} + \mu \cdot \min(v_{\text{max}}, \|\mathbf{v}\|_2 \cdot \alpha) \right] \cdot M_{\text{shoot}}(d_1,d_2)
\]
Here, \(\hat{\mathbf{v}} = \mathbf{v}_{\text{robot}} / (\|\mathbf{v}_{\text{robot}}\|_2 + \varepsilon)\) normalizes the robot's velocity direction, and \(\hat{\mathbf{v}}_{\text{ball-goal}}\) is the unit vector from the ball to the goal. The parameter \(\mu\) is the speed reward coefficient, \(v_{\text{max}}\) is the maximum speed reward, and \(\alpha\) is the speed amplification factor. The phase mask \(M_{\text{shoot}}(d_1,d_2)\) is defined as:
\[
M_{\text{shoot}}(d_1,d_2) = \begin{cases}
1 & \text{if } 0 < d_1 \leq u_2, \, d_2 < u_3 \\
0 & \text{otherwise}
\end{cases}
\]
This reward encourages alignment with the goal and rewards higher speeds for more powerful shots.

\subsubsection{Command Smoothness Regularization}
To prevent erratic behavior and ensure smooth transitions between commands, a regularization term is introduced:
\[
r_{\text{delta}} = -\zeta_1 \cdot \max(0, \|\Delta \mathbf{v}\|_2 - \delta_{\text{max}}) - \zeta_2 \cdot \mathbf{1}[\|\Delta \mathbf{v}\|_2 < \delta_{\text{min}}]
\]
Here, \(\zeta_1\) penalizes excessive velocity commands beyond a maximum threshold \(\delta_{\text{max}}\), while \(\zeta_2\) penalizes too small commands, enforcing controlled actions with meaningful impact.

This multistage reward design ensures phase consistency by activating only one stage at a time, preventing conflicting objectives. The phases build on each other, from basic approach to complex shooting, while continuous distance thresholds prevent reward discontinuities. Each stage encourages distinct motor patterns that are appropriate for the soccer task context.

\begin{table*}[htbp]
  \centering
  \caption{\textbf{Summary of Multi-Stage and Standard Reward Functions}. The total reward is a weighted sum of task-specific rewards (active only in specific phases) and standard regularization rewards (active globally).}
  \label{tab:reward_summary}
  \resizebox{\textwidth}{!}{%
  \begin{tabular}{l l l l}
    \toprule 
    \textbf{Category} & \textbf{Component} & \textbf{Mathematical Formulation} & \textbf{Description / Condition} \\
    \midrule 
    \multicolumn{4}{l}{\textit{\textbf{Task-Specific Rewards (Phase-Dependent)}}} \\
    \midrule
    Approach & $r_{\text{approach}}$ & $(\mathbf{v}_{\text{robot}} \cdot \hat{\mathbf{v}}_{\text{robot-ball}}) \cdot M_{\text{approach}}$ & Encourage moving towards the ball ($d > u_1$) \\
    Alignment & $r_{\text{alignment}}$ & $[\cos(\theta_{\text{rbg}}) - \beta |\Delta \omega_{\text{err}}|] \cdot M_{\text{alignment}}$ & Align robot-ball-goal vector ($u_2 < d \le u_1$) \\
    Dribble & $r_{\text{dribble}}$ & $(\mathbf{v}_{\text{robot}} \cdot \hat{\mathbf{v}}_{\text{ball-goal}}) \cdot M_{\text{dribble}}$ & Guide ball towards goal gently ($d_1 \le u_2, d_2 > u_3$) \\
    Shoot & $r_{\text{shoot}}$ & $[\text{Align}_{\text{goal}} + \mu \cdot \min(v_{\text{max}}, v_{\text{shoot}})] \cdot M_{\text{shoot}}$ & Maximize kick velocity and accuracy ($d_1 \le u_2, d_2 < u_3$) \\
    \midrule
    \multicolumn{4}{l}{\textit{\textbf{Standard \& Regularization Rewards (Global)}}} \\
    \midrule
    Smoothness & $r_{\text{delta}}$ & $-\zeta_1 \|\Delta \mathbf{v}\| - \zeta_2 \mathbb{I}(\|\Delta \mathbf{v}\| < \delta_{\text{min}})$ & Penalize erratic or trivial command changes \\
    Survival & $r_{\text{survival}}$ & $1.0 \text{ if } h_{\text{root}} > h_{\text{thresh}} \text{ else } 0$ & Encourage the robot to remain upright \\
    Orientation & $r_{\text{orient}}$ & $-\| \mathbf{g}_{\text{proj}} - \mathbf{g}_{\text{vertical}} \|^2$ & Penalize base tilt (pitch/roll deviation) \\
    Energy & $r_{\text{energy}}$ & $-\sum \|\boldsymbol{\tau}\|^2$ & Penalize excessive joint torque consumption \\
    \bottomrule 
  \end{tabular}}
\end{table*}
The comprehensive definition of all reward components, including the standard survival and regularization terms used to ensure training stability, is detailed in Table~\ref{tab:reward_summary}.
\subsection{Visual Perception}
\label{sec:visual_perception}
Real-time ball detection is achieved using the YOLOv8 object detection algorithm, which operates on the robot's onboard GPU. The detection pipeline processes RGB images at a frequency of 30 Hz, outputting bounding boxes for the detected ball. These 2D bounding box coordinates are then converted into 3D positions by incorporating depth information and camera calibration parameters.

The ball's position in the robot frame is computed using the following transformation:
\[
\mathbf{p}_{\text{ball}} = \mathbf{R}_{\text{cam}}^{\text{robot}} \cdot \mathbf{K}^{-1} \cdot [x \cdot d, \, y \cdot d, \, d]^T + \mathbf{t}_{\text{cam}}^{\text{robot}}
\]
In this equation, \(\mathbf{K}\) represents the camera's intrinsic matrix, and \((x, y, d)\) are the detected 2D ball coordinates, with \(d\) denoting the depth (distance from the camera to the ball). The terms \(\mathbf{R}_{\text{cam}}^{\text{robot}}\) and \(\mathbf{t}_{\text{cam}}^{\text{robot}}\) represent the rotation matrix and translation vector, respectively, which transform the ball's position from the camera frame to the robot frame.

\section{Experiment}
\subsection{Experimental Setup}
\begin{figure}[htbp]
\centering
\includegraphics[width=1.0\linewidth]{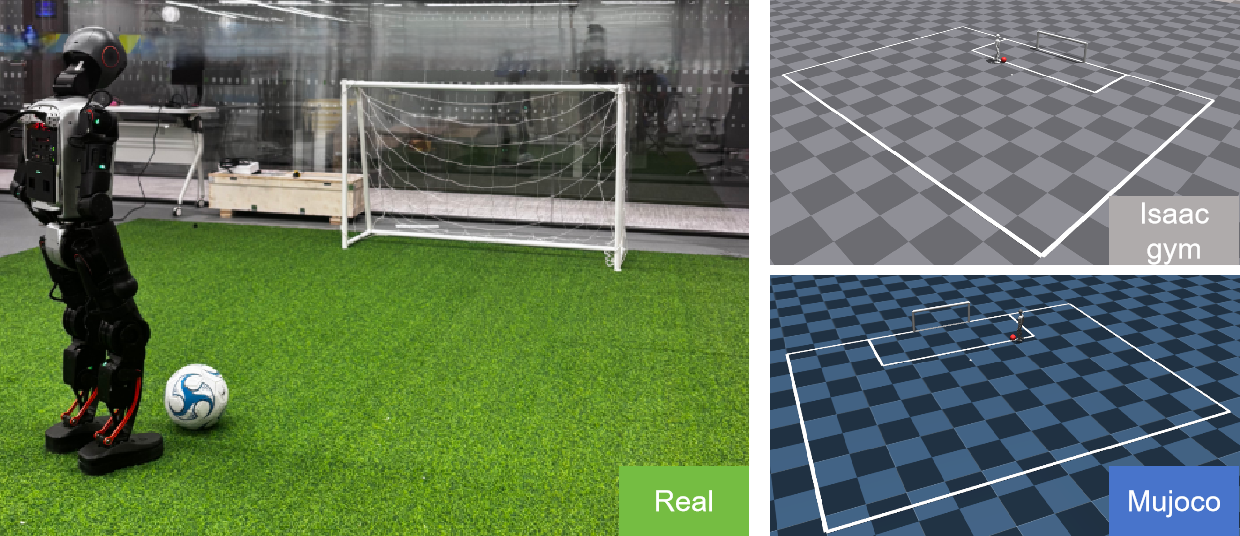}
\caption{\label{fig:kick_environment}\textbf{Simulation and Real-World Environments}. The environments for the football robot's kicking task, including both simulated and real-world setups, are used for algorithm development, verification, and deployment.}
\end{figure}

The HierKick framework is evaluated across three distinct environments to assess its effectiveness and transferability: Isaac Gym, MuJoCo, and real-world deployment on the Booster T1 humanoid robot. Isaac Gym is used for high-fidelity physics simulation, MuJoCo serves for rapid prototyping, and real-world deployment is conducted on the Booster T1 robot. All experimental environments, as illustrated in Fig.~\ref{fig:kick_environment}, feature standardized soccer fields equipped with goals and regulation-size soccer balls.

\subsubsection{Environments Overview}
\textit{Isaac Gym} is the primary simulation platform, enabling GPU-parallel simulations for data-efficient Proximal Policy Optimization (PPO) training, with extensive domain randomization to improve generalization. \textit{MuJoCo} serves as an independent physics engine to assess the robustness of the model against variations in physical modeling. \textit{Real-World Platform} utilizes the Booster T1 humanoid robot on artificial turf, equipped with a small goal and a standard soccer ball (see Fig.~\ref{fig:kick_environment}).

\subsubsection{Perception and Control Stack}
For perception, YOLOv8 is deployed on the real robot, running at approximately 10 Hz to detect the ball and goal. The camera is calibrated both intrinsically and extrinsically, with detected 2D bounding boxes converted into 3D coordinates using depth information and camera calibration parameters (see Sect.~\ref{sec:visual_perception}). The control stack consists of two main modules: the Coach policy and the Motion policy. These modules are integrated with pre-trained kick skills and operate at 50 Hz to track velocity commands and generate 12-degree-of-freedom joint targets. The end-to-end latency from perception to action is approximately 20 ms at 50 Hz.

\subsubsection{Initialization and Randomization}
Ball and robot poses, as well as robot headings, are randomized at the beginning of each trial. In simulation, additional randomizations include ground friction, ball mass, radius and bounciness, camera extrinsics, sensor noise, and command delay jitter to enhance the robustness and generalization of the model.
The specific hyperparameters for the PPO algorithm and the network architecture used in our experiments are listed in Table~\ref{tab:hyperparameters}.
\begin{table}[htbp]
  \centering
  \caption{\textbf{Hyperparameters and Configuration Details}.}
  \label{tab:hyperparameters}
  \small  
  \renewcommand{\arraystretch}{1.1}
  \begin{tabular}{l l c}
    \toprule
    \textbf{Category} & \textbf{Parameter} & \textbf{Value} \\
    \midrule
    \multirow{8}{*}{\textbf{PPO Config}} 
      & Discount Factor ($\gamma$) & 0.99 \\
      & GAE Parameter ($\lambda$) & 0.95 \\
      & Clip Range ($\epsilon$) & 0.2 \\
      & KL Target & 0.01 \\
      & Learning Rate & $3 \times 10^{-4}$ \\
      & Batch Size & 4096 \\
      & Mini-batch Size & 1024 \\
      & Optimization Epochs & 5 \\
    \midrule
    \multirow{4}{*}{\textbf{Env. \& Control}} 
      & High-Level Freq. & 5 Hz \\
      & Low-Level Freq. & 50 Hz \\
      & Max Episode Length & 20 s \\
      & Latency (Sim) & $\sim$20 ms \\
    \midrule
    \multirow{3}{*}{\textbf{Network}} 
      & Actor MLP & $[512, 256, 128]$ \\
      & Critic MLP & $[512, 256, 128]$ \\
      & Activation & ELU \\
    \bottomrule
  \end{tabular}
\end{table}
\subsection{Evaluation and Metrics}
Each trial begins with a randomized kickoff configuration and must be completed within 20 seconds. A trial is deemed successful if two conditions are met: first, the ball must fully cross the goal line before the 20-second time limit. Second, the robot must remain upright throughout the trial. Two primary metrics are used to quantify performance.

\subsubsection{Task Success Rate}
\begin{figure}[htbp]
    \centering
    \includegraphics[width=0.7\linewidth]{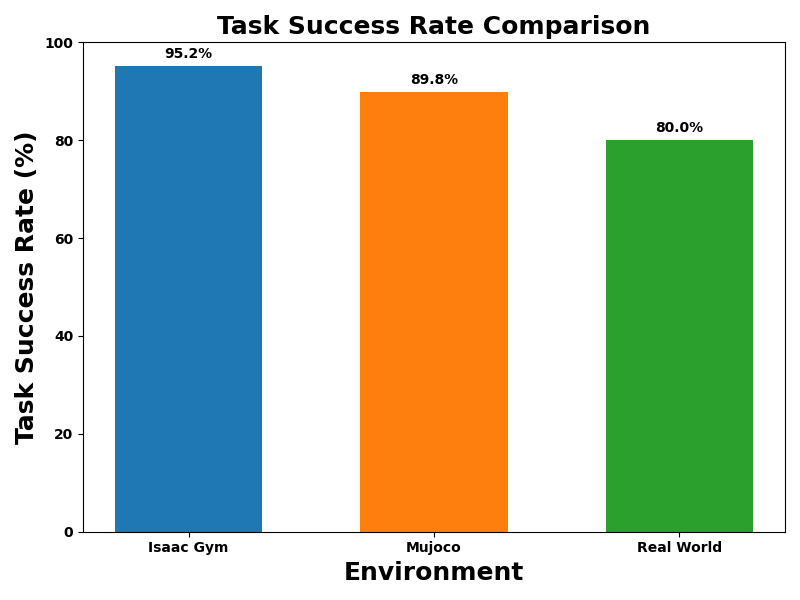}
    \caption{\textbf{Success Rate Comparison}. This graph denotes the comparison of task success rates across simulation environments and real-world deployment.}
    \label{fig:success_rate_comparison}
\end{figure}

The task success rate is calculated for Isaac Gym, MuJoCo, and the real robot (see Fig.~\ref{fig:success_rate_comparison}), and it represents the proportion of successful trials within each environment. To verify the necessity of key observations and the rationality of output design in HierKick, two ablative experiments are conducted. One is HierKick Without $\mathbf{d}_{\text{ball}}$, $\mathbf{d}_{\text{goal}}$, where the strategy is stripped of the observation information $\mathbf{d}_{\text{ball}}$ and $\mathbf{d}_{\text{goal}}$, forcing it to make decisions without these critical positional cues. The other is HierKick ($\mathbf{c}_{\text{prev}}\!\to\!\mathbf{v}_{\text{robot-ball}}$), where the original acceleration-related command output $\mathbf{c}_{\text{prev}}$ of the strategy is replaced with $\mathbf{v}_{\text{robot-ball}}$, to test the impact of output type on task performance. The success rates of these ablative variants, along with the original HierKick and End-to-End methods, are summarized in Table~\ref{tab:component_impact}.

\begin{table}[htbp]
  \centering
  \caption{\textbf{Impact of Component Changes and HierKick Approaches (vs. End-to-End) on Model Success Rate}}
  \label{tab:component_impact}
  \begin{tabular}{l c}
    \toprule 
    \textbf{Compared Methods} & \textbf{Success Rate} \\ 
    \midrule 
    HierKick & \textbf{95.2\%} \\ 
    HierKick w/o $\mathbf{d}_{\text{ball}}$, $\mathbf{d}_{\text{goal}}$ & 23.2\% \\
    HierKick ($\mathbf{c}_{\text{prev}}\!\to\!\mathbf{v}_{\text{robot-ball}}$) & 8.2\% \\
    End-to-End & 25.6\% \\
    \bottomrule 
  \end{tabular}
\end{table}

\subsubsection{Kick Distance}
The kick distance is defined as the distance between the final ball position and the goal line, and it is recorded only for successful trials. To ensure statistical reliability, each simulated configuration is evaluated over twenty thousand randomized trials. For the real robot, testing is conducted across multiple sessions with varied lighting conditions and different initializations.

\subsection{Results Across Platforms}
Fig.~\ref{fig:success_rate_comparison} summarizes the cross-platform performance of HierKick. For Isaac Gym, the success rate is 95.2\%, benefiting from high-fidelity simulation with minimal physical modeling errors. For MuJoCo, the success rate is 89.8\%, reflecting a 5.4\% drop from Isaac Gym. This reduction primarily stems from differences in contact and friction modeling, as well as engine-specific integration issues, which slightly destabilize dribbling and final-step alignment. For the real robot, the success rate is 80.0\%, a further 9.8\% drop from MuJoCo. With the inclusion of \(\mathbf{c}_{\text{prev}}\), the coach policy is able to filter occasional misdetections without destabilizing the system's commands. The performance degradation in the real robot can be attributed to real-world uncertainties, which contribute to the sim-to-real gap. These gaps arise from factors such as local turf friction, ball rebound differences, and the robot's interaction with the environment. Domain randomization mitigates many of these issues, but extreme turf patches continue to affect dribbling performance. Common failure modes contributing to the performance drop include partial ball occlusion delaying alignment, low-friction areas causing foot slip during the final step, and glare introducing depth bias. Despite these challenges, the 80.0\% success rate demonstrates strong sim-to-real transferability of the hierarchical policy.

\begin{figure}[htbp]
    \centering
    \includegraphics[width=\linewidth]{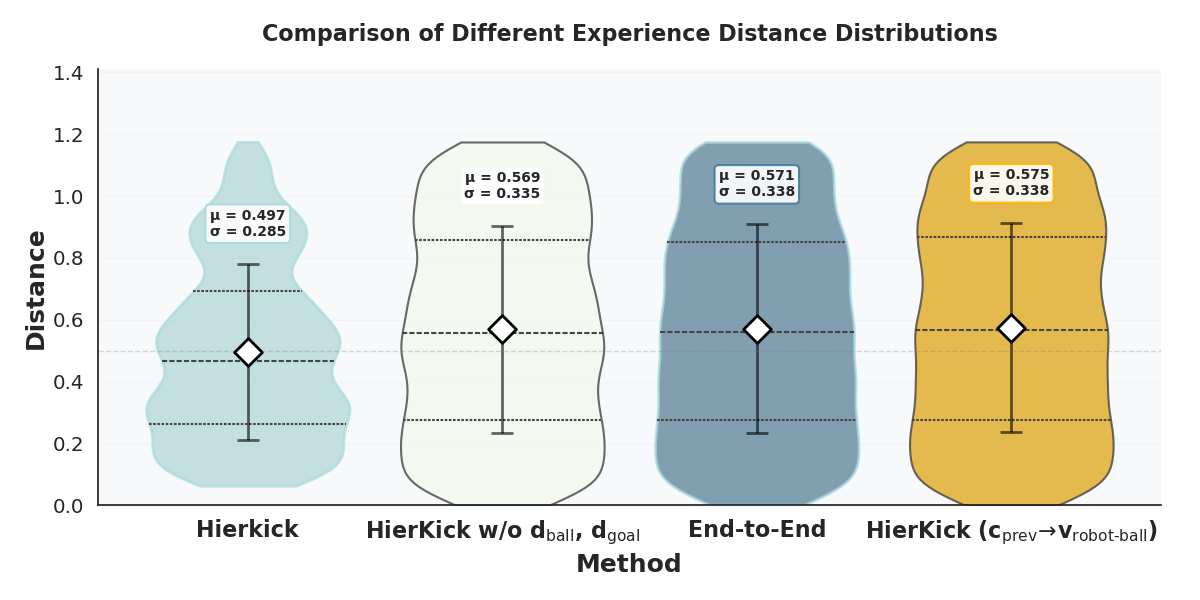}
    \caption{\textbf{Horizontal Distance to Goal Center}. Kick distance distributions under three conditions. HierKick achieves the smallest mean and most concentrated distribution, whereas Remove- and HierKick ($\mathbf{c}_{\text{prev}}\!\to\!\mathbf{v}_{\text{robot-ball}}$) experiment exhibit larger means. White diamonds denote means; error bars indicate one standard deviation. The dashed line at 0.5 m marks a reference threshold. Each condition has 5280 samples.}
    \label{fig:holder}
\end{figure}

\subsection{Kick Distance Analysis}
We analyze the kick distance (horizontal distance between the ball's landing point and goal center) distribution across four configurations: complete HierKick, HierKick w/o \(\mathbf{d}_{\text{ball}}\), \(\mathbf{d}_{\text{goal}}\), End-to-End, and HierKick (\(\mathbf{c}_{\text{prev}}\!\to\!\mathbf{v}_{\text{robot-ball}}\)). Fig.~\ref{fig:holder} illustrates this distribution, with trends consistent with the ablation results in Table~\ref{tab:component_impact}. The complete HierKick shows the most concentrated distribution---smallest variance and lowest mean kick distance---indicating stable, precise kicking, thanks to its comprehensive observation components that enable accurate alignment and speed control. In contrast, End-to-End, HierKick (\(\mathbf{c}_{\text{prev}}\!\to\!\mathbf{v}_{\text{robot-ball}}\)), and HierKick w/o \(\mathbf{d}_{\text{ball}}\), \(\mathbf{d}_{\text{goal}}\) exhibit larger mean distances and wider distributions. This confirms that End-to-End and HierKick variants (omitting critical perceptual components or altering command history) lead to systematic misalignment or insufficient final strike speed---consistent with their low success rates in Table~\ref{tab:component_impact}.

\subsection{Reward Comparison with different Method}
\begin{figure}[htbp]
    \centering
    \includegraphics[width=0.9\linewidth]{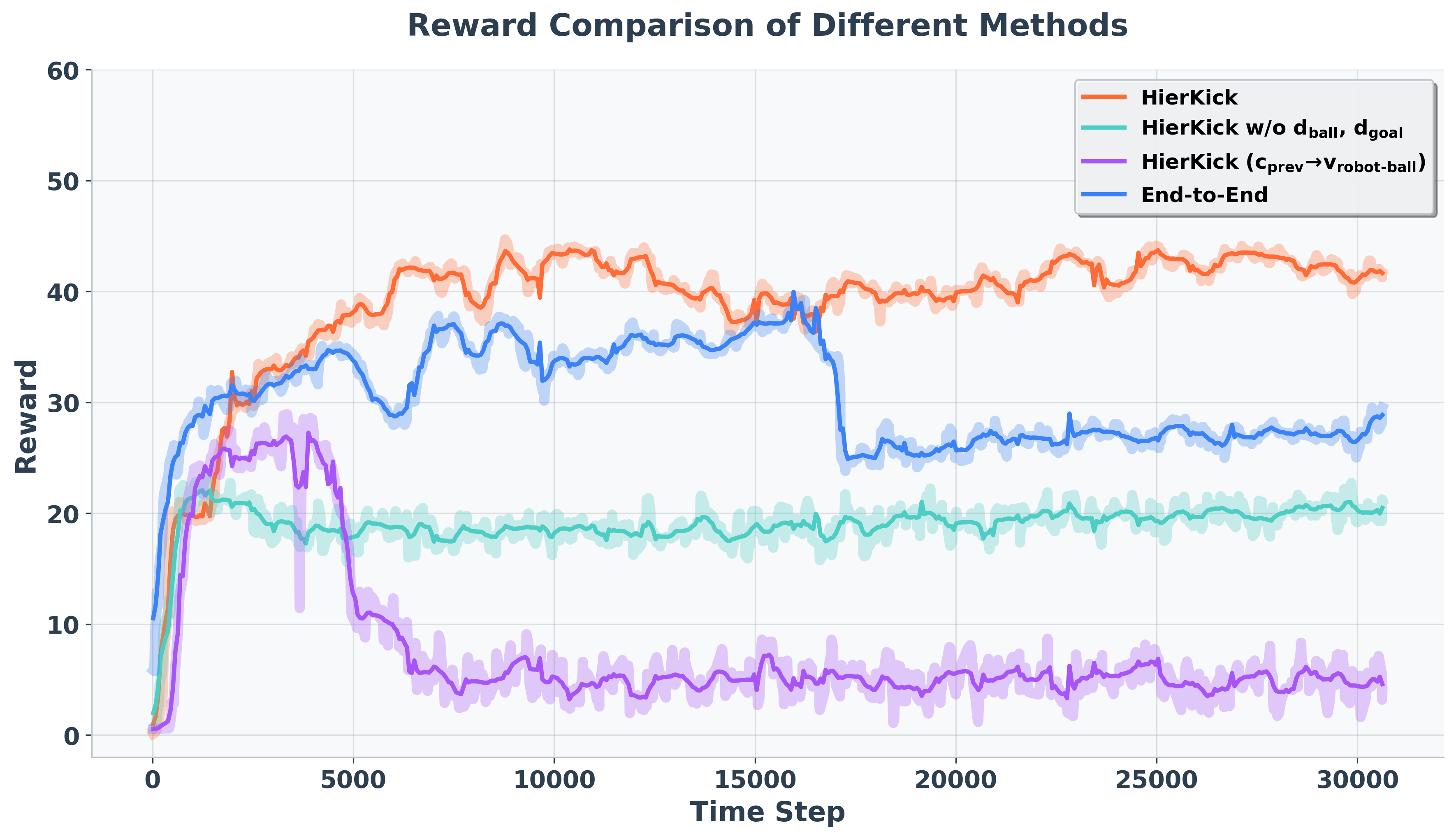}
    \caption{\textbf{Reward Comparison}. Comparison of Reward Evolution Between End-to-End and HierKick.}
    \label{fig:placeholder}
\end{figure}

Fig.~\ref{fig:placeholder} compares the reward evolution between different methods during training. This figure compares the rewards of four methods during training: HierKick, HierKick w/o \(\mathbf{d}_{\text{ball}}\), \(\mathbf{d}_{\text{goal}}\), HierKick (\(\mathbf{c}_{\text{prev}}\!\to\!\mathbf{v}_{\text{robot-ball}}\)), and End-to-End. HierKick achieves the highest and most stable rewards overall, while the other methods have relatively lower rewards, which verifies the superiority of our HierKick method.

\section{CONCLUSION}
We have presented HierKick, a dual-frequency hierarchical reinforcement learning (RL) framework for vision-guided humanoid soccer robots. The framework employs a 5 Hz coach policy that issues velocity increments, while a 50 Hz pre-trained controller executes these commands using YOLOv8 for perception, closing the loop. This frequency-separated design strikes a balance between long-horizon tactical planning and precise, fast joint control during the approach, alignment, dribbling, and shooting phases. Multi-stage rewards and domain randomization ensure stable training and sim-to-real transfer, with approximately 20 ms end-to-end latency. Ablation studies show that ball-relative geometry and command history are crucial components for success. The reuse of pre-trained skills reduces training time and improves convergence. The hierarchical approach leads to more precise, lower-variance kick distances. However, limitations include challenges such as occlusions, turf variability, last-contact uncertainty, fixed phase thresholds/decision rates, and modest perception fidelity.

\bibliographystyle{unsrt}
\bibliography{ref}
\end{document}